\documentclass[runningheads]{llncs}
\usepackage[T1]{fontenc}
\usepackage{graphicx}
\usepackage{amsmath, % for math equations
            amssymb, % for math and greek symbols
            mathtools} % for math equations 
\usepackage{nicematrix}

\newcommand\norm[1]{\left\lVert #1 \right\rVert}

\begin{document}
\title{Generative modeling of Sparse Approximate Inverse Preconditioners}
\author{Mou Li\inst{1}\orcidID{0009-0009-0216-9292} \and
He Wang\inst{2}\orcidID{0000-0002-2281-5679} \and
Peter K.\ Jimack\inst{3}\orcidID{0000-0001-9463-7595}}
\authorrunning{M. Li et al.}
\institute{University of Leeds, Leeds LS2 9JT, UK \\
\email{scmli@leeds.ac.uk}\and
University College London, London WC1E 6BT, UK\\
\email{he\_wang@ucl.ac.uk}\\
\url{http://drhewang.com}\and
University of Leeds, Leeds LS2 9JT, UK\\
\email{p.k.jimack@leeds.ac.uk}\\
}
\maketitle              % typeset the header of the contribution
\begin{abstract}
We present a new deep learning paradigm for the generation of sparse approximate inverse (SPAI) preconditioners for matrix systems arising from the mesh-based discretization of elliptic differential operators. Our approach is based upon the observation that matrices generated in this manner are not arbitrary, but inherit properties from differential operators that they discretize. Consequently, we seek to represent a learnable distribution of high-performance preconditioners from a low-dimensional subspace through a carefully-designed autoencoder, which is able to generate SPAI preconditioners for these systems. The concept has been implemented on a variety of finite element discretizations of second- and fourth-order elliptic partial differential equations with highly promising results.

\keywords{Deep learning \and Sparse matrices  \and Preconditioning \and Elliptic partial differential equations \and Finite element methods.}
\end{abstract}
\section{Introduction}
\label{sec:intro}

Finding the solution of systems of linear algebraic equations,
\begin{equation}
    Ax = b \;, \label{LinearSystem}
\end{equation}
has been a core topic in the field of scientific computation for many decades. Such systems arise naturally in a wide range of algorithms and applications, including from the discretization of systems of partial differential equations (PDEs). Not only are the equations~(\ref{LinearSystem}) the main product of the discretization process for linear PDEs but such systems must be solved during every nonlinear iteration when solving nonlinear PDE systems~\cite{benzi_preconditioning_2002}. Furthermore, this step is often the most computationally expensive in any given numerical simulation.

Direct methods, based on the factorisation of the matrix $A$ into easily invertible triangular matrices are designed to be highly robust. However they generally scale as $O(n^3)$ as the system size, $n$, increases. Hence as $n$ approaches values in the millions the computational cost and memory requirements quickly become unacceptable. Fortunately, when the matrix $A$ is sparse, it is possible to optimize the factorization process to minimize the fill-in and avoid storing (or computing with) the zero entries in $A$~\cite{li_splu05}. Such sparse direct methods significantly reduce the computational complexity but can still suffer from unacceptable memory and CPU requirements when $n$ becomes sufficiently large. Consequently iterative solution methods are a popular choice for large sparse systems~(\ref{LinearSystem}).

Many iterative methods have been developed as an alternative approach to direct methods, offering a flexible level of accuracy as the trade-off for memory storage and computational cost. The Conjugate Gradient (CG) algorithm is arguably the most well-developed and efficient iterative method for solving large, sparse, symmetric positive definite (SPD) systems, whilst other Krylov-subspace algorithms are widely-used for indefinite or non-symmetric systems~\cite{Golub2013}. In the SPD case, the convergence of the CG method is dependent on the distribution of the eigenvalues of the stiffness matrix $A$~\cite{tadmor_review_2012}. Assuming that these eigenvalues are widely spread along the real line, the CG algorithm will suffer from slow convergence when the problem is ill-conditioned. That is, the condition number $\kappa(A) \gg 1$, which is the ratio of the maximum, $\lambda_{\max}(A)$, over the minimum, $\lambda_{\min}(A)$, eigenvalues of $A$. Consequently, a longstanding topic of research has been the development of  preconditioning techniques to reduce $\kappa(A)$ (strictly $\kappa (AP)$) via a transformation of the stiffness matrix $A$ through multiplication with a carefully selected non-singular matrix $P$ on both sides of the equation: $APx = bP$.

Selection of a ``good'' preconditioner, $P$ should take account of the following two requirements:
\begin{enumerate}
\item It should be easy to construct and cheap to implement (i.e. to solve $Px=y$ for a given vector $y$).
\item It should significantly reduce the condition number of the modified problem, $\kappa(AP)$.
\end{enumerate}
In general these requirements are in conflict with each other, so frequently we seek to use prior knowledge about the linear system in order to manage the trade-off between point one and two. If little is known about $A$ then one of the simplest choices for $P$ is the diagonal of $A$ (frequently referred to as Jacobi preconditioning~\cite{Golub2013}), which clearly satisfies the first requirement above but typically satisfies the second requirement less well. More complex preconditioners, which aim to do better on the second requirement, include incomplete factorization approaches such as incomplete Cholesky (IC) and incomplete LU (ILU) decompositions~\cite{Chan1997,Golub2013}, multigrid (MG) and algebraic multigrid (AMG) approaches~\cite{Bramble1993,Van_AMG_2002}, domain decomposition (DD) methods~\cite{Quarteroni1999}, and sparse approximate inverse (SPAI) methods~\cite{AnztHartwig_ISPAI_2018} which are the main topic of this paper. In recent developments there are also some attempts to utilise data-driven methods to generate preconditioners~\cite{Azulay2021,hausner_neural_2024,kopanicakova_deeponet_2024,li_learning_2023,lu_deeponet_2021,sappl_deep_2019}.

If the exact inverse of $A$ were known then setting $P=A^{-1}$ would perfectly satisfy condition 2 above. However, the cost of constructing and implementing this $P$ would mean that condition 1 would fail to be satisfied. Sparse approximate inverse (SPAI) methods seek to balance these requirements more evenly by constructing a cheaper approximation to $A^{-1}$ ($P\approx A^{-1}$), where $P$ is a sparse matrix so as to ensure that its application, $y=Px$, has a cost of $O(n)$ rather than $O(n^2)$. Designing a high performance preconditioner by the conventional numerical approaches mentioned above requires prior knowledge and past experience of specific families of PDE systems. The final product also varies with chosen parameters, this suggests that there is a distribution of high-performance preconditioners for $A$, which naturally justifies using generative models. In this paper, we propose a deep learning based generative model for constructing a SPAI preconditioner, $P$, for a given SPD matrix $A$ arising from the finite element discretization of a self-adjoint PDE or system. Our method uses a conditioned variational auto-encoder to map the conditioned distribution of $p(A^{-1} | A)$ into a lower dimensional latent space. After training, it is able to generate high-performance preconditioners for SPD matrices arising from the discretization of unseen self-adjoint problems under the same mesh density.
    
Whilst this is not the first research to propose the use of machine learning (ML) techniques to generate SPAI preconditioners (see \cite{sappl_deep_2019} for example), we believe that this is the first to consider the use of a graph representation of the sparse matrix $A$ and the first to consider modeling the distribution of $A$ and $A^{-1}$ for such a task. This permits the representation of $A$ in a smaller latent space through our use of a variational auto-encoder. Consequently, since our model only cares about the latent data distribution of the inverse of the stiffness matrix A, it is easily generalisable to other self-adjoint problems. Furthermore, we propose a controllable sparsity pattern for the preconditioner to allow a trade-off between the performance and the computational cost.

\section{Related Work}
\label{sec:related work}
In this section we provide a brief overview of the most common approaches to preconditioning that are in widespread use in numerical codes today. This is then followed by a short subsection on existing research into the use of data-driven methods for generating preconditioners. As part of that section we also highlight some of the proposed techniques for using ML to solve PDE systems since these could also be used to motivate new preconditioners (i.e. by applying as a preconditioner rather than a solver).

\subsection{Numerical preconditioning}
\label{subsec:numerical pre}

Researchers have studied different methods for generating effective preconditioners over many decades, leading to a vast body of work on this topic~\cite{wathen_preconditioning_2015}.
For the purposes of this paper we restrict our attention to sparse SPD matrices, $A$, arising from mesh-based discretizatons of self-adjoint PDEs. As already noted, the simplest approach that is in widespread use is
{Jacobi preconditioning}, where $P = \text{diag}(A)^{-1}$. This is easy to construct and cheap to implement, but not sufficiently effective for many important problems~\cite{li_learning_2023}.
Consequently more sophisticated techniques are frequently required, such as those based upon incomplete factorizations of $A$. Incomplete Cholesky (IC) factorization is most appropriate for SPD matrices, where $A \approx L L^T$. The sparsity pattern of the lower triangular matrix $L$ can be chosen to be identical to that of the lower triangle of $A$ or can be slightly more generous based upon a drop tolerance that is used to decide whether to allow any "fill-in" during the factorization process  \cite{wathen_preconditioning_2015}. Note that this technique requires a forward and a backward substitution in order to apply the preconditioner, which is not generally well-suited to efficient parallel implementation. Furthermore, this approach requires careful design to balance the computational cost and accuracy associated with different levels of fill-in (complete fill-in leads to perfect factorization, which means that $\kappa(AP)=1$ but at prohibitive cost, whereas insufficient fill-in can lead to a sub-optimal condition number).

The other class of preconditioners that we discuss in detail here are sparse approximate inverse (SPAI) methods, which are the main topic of this paper. This variant evaluates the preconditioner $P$ to be a sparse approximation to the inverse of the stiffness matrix $A$. Frobenius norm minimisation and incomplete bi-conjugation are the two most widely implemented frameworks to compute the SPAI preconditioners \cite{benzi_preconditioning_2002,wathen_preconditioning_2015}. One major difficulty of SPAI preconditioning is the choice of the sparsity pattern. Since the inverse of an irreducible sparse matrix is proven to be a structurally dense matrix \cite{duff_sparsity_1988}, when the sparsity pattern is predefined, such preconditioners may not work well if there exist entries with large magnitude outside the defined pattern. Attempts to address this by automatically capturing a pseudo-optimal sparsity pattern include the use of a drop tolerance \cite{chow_approximate_1998} or of a ``profitability factor'' via a residual reduction process \cite{grote_parallel_1997,huckle_factorized_2003}.
In this work we consider only predefined sparsity patterns based around the sparsity of the family of matrices, $A$, being considered.

\subsection{Data driven preconditioning}
\label{subsec:DL pre}
In recent years a number of approaches have been proposed for the direct solution of systems of PDEs using both supervised and unsupervised ML methods. Noteworthy, and highly influential, early examples include the deep Ritz method~\cite{Wienan_E2018},
the deep Galerkin method~\cite{sirignano_dgm_2018}
and physics-informed neural networks (PINNs)~\cite{raissi_physics-informed_2019,raissi_physics_2017-1,raissi_physics_2017}.
The latter approach has led to a substantial, and rapidly increasing, body of research into the unsupervised learning of PDE solutions and related inverse problems. However, as forward solvers, PINNs are not generally competitive with classical numerical approaches based upon efficient preconditioning. It is for this reason, and inspired by the successes of PINNs, that we seek to utilise the power of machine learning to generate high-performance preconditioners (as opposed to complete solvers).

There is relatively little prior work on the use of machine learning to develop preconditioners for use within conventional numerical algorithms. The first attempt to generate a SPAI preconditioner appears in~\cite{sappl_deep_2019}, where a convolutional neural network (CNN) is used to derive an approximate triangular factorization of the inverse matrix. More recent research, such as \cite{hausner_neural_2024,li_learning_2023}
has focused on incomplete LU and Cholesky preconditioners, with the former targeting the non-self-adjoint Navier-Stokes equations and the latter two considering SPD systems: the work of~\cite{li_learning_2023} being most similar to that considered in this paper due to their use of a GNN. Other approaches that have been taken to develop novel preconditioners include~\cite{Azulay2021},
which mimics a multigrid approach through a combination of a CNN-based smoother and coarse-grid solvers, and~\cite{kopanicakova_deeponet_2024},
which builds a preconditioner directly upon a DeepONet approximation of the operator that represents the solution of the underlying PDE~\cite{lu_deeponet_2021}.

\section{Methodology}
\label{sec:method}
In this initial investigation we focus on distributions of sparse matrices, $A$, arising from the finite element discretization of elliptic PDEs: the precise sparsity pattern depending upon the differential operator, the finite element spaces used, and the mesh on which the discretization occurs. Even though the inverse matrices, $A^{-1} \in \mathbb{R}^{n \times n}$ are generally dense in structure, other preconditioning methods are able to approximate cheaper versions of $A^{-1}$, which led us to assume that there exists a learnable distribution of high performance preconditioners within a lower dimensional subspace of $\mathbb{R}^{n \times n}$ with a prescribed sparsity pattern. We then propose a graph-conditioned variational autoencoder (GCVAE) architecture which is able to generate such SPAI preconditioners for these SPD linear systems.

\subsection{Problem setup}
\label{subsec:problem setup}
Inspired by the traditional Frobenius norm minimization approach of SPAI preconditioning, our starting point is the following assumption:
\begin{equation}
\label{eq:SPAI}
\forall A \in S_A\;\;\; \exists R \in S_M : \norm{I - R^{T} A R}_F \approx 0   \;,
\end{equation}
where $S_A$ is our set of possible sparse SPD matrices and $S_M \subset \mathbb{R}^{n \times n}$ with a prescribed sparsity pattern defined by a selected mask, $M$.
Let us consider the dataset ${\cal A} = \{A^{(i)}\}_{i=1}^N$ having $N$ independently and identically distributed samples and drawn from an unknown distribution $p({\cal A})$, generated from a given family of linear PDE problems. Let ${\cal A}^{-1} = g({\cal A})$ where $g(\cdot)$ is the inverse matrix transformation function. We assume that the conditional distribution of $p({\cal A}^{-1} | {\cal A})$ lies within a parametric family of distributions, such as Gaussian, on a lower-dimensional latent space such that $p_\theta (z | {\cal A},{\cal A}^{-1}) = \mathcal{N}(\mu,\sigma^2)$, where $\theta$ represents the learnable parameters of the neural networks. We also assume that for every given $A^{(i)}$ there exists a set of high-performance preconditioners ${\cal R}^{(i)}$ which satisfies eq.~\ref{eq:SPAI} with a given sparsity pattern $M$ and ${\cal R} \sim p_{\theta} (z | {\cal A},{\cal A}^{-1})$. Then the SPAI preconditioners $\cal R$ can be generated from the generative distribution $p_\theta ({\cal R} | z, \cal A)$ conditioned on input $\cal A$, the prescribed sparsity pattern of $\cal R$, and latent variable $z \sim \mathcal{N}(\mu,\sigma^2)$.

The application of variational auto-encoders (VAEs) has shown a great potential in learning the probability distribution of a given dataset \cite{kingma_auto-encoding_2014} and, when coupled with a condition, can ensure the robustness of the generative model \cite{mirza_conditional_2014}. VAEs have also shown great potential in modeling the distribution of the utility matrix for a recommender system~\cite{ali_variational_2021} and for a gene expression matrix \cite{lukassen_gene_2020}, both based upon large and sparse matrices. Therefore, we propose a conditional VAE generative model to approximate SPAI preconditioners.\\

\subsection{Architecture design}
\label{subsec:architecture}
As noted above, the type of architecture we now consider is a conditioned variational auto-encoder (cVAE) \cite{gu_diversity-promoting_2021}. The encoder part consists of two different encoders, which we now discuss in turn.

\begin{figure}
\includegraphics[width=\textwidth]{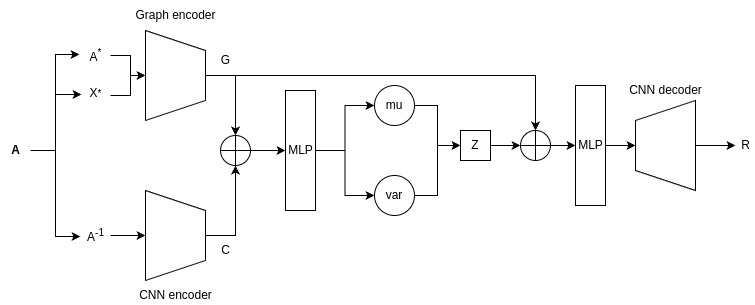}
\caption{Schematic diagram of the proposed graph-conditioned variational autoencoder} 
\label{fig:Model schematic diagram}
\end{figure}

As shown in Figure~\ref{fig:Model schematic diagram}, the first is a conditional part (a graph encoder) that consists of several layers of a graph neural network (GNN). The input data $A$ is a sparse matrix which is represented as a graph, where the diagonal entries and off-diagonal entries are the nodes and edges of the graph respectively. GNNs have two clear advantages over CNNs under our setting: 1) a GNN can perform the feature mapping with a full ``view'' of the input data through its message passing mechanism, 2) a GNN is only interested in the entries that are non-zero therefore it is a sparse implementation, which significantly reduces the memory requirement and the training cost.
The graph VAE is first proposed by \cite{kipf_variational_2016}, where it is used to learn latent representations of unweighted undirected graphs with multidimensional node features for link prediction in citation networks. In our case, the system matrix $A$ can be treated as a weighted and directed graph with single dimensional node features. Therefore we replace the graph convolutional layer (GCN) with the graph attention layer (GATv2) \cite{brody_how_2022}. This graph encoder takes $A^*$ and $X^*$ as the input, where $A^*$ is the self-looped adjacency matrix of $A$ and $X^*$ is the node feature matrix (where $X^* \in \mathbb{R}^{n\times1}$ and $n$ is the number of nodes).

The second encoder is chosen to be a CNN encoder since it takes $A^{-1}$, the exact inverse of the system matrix $A$, as the input. CNNs are known to be powerful and efficient feature extractors for image data \cite{goodfellow_deep_2016}, where grey-scale images are fundamentally dense matrices. During training, we assume the latent distribution is a Gaussian, i.e. $z \sim \mathcal{N}(\mu,\sigma^2)$, both encoders’ outputs are concatenated and then passed through a multilayer perceptron (MLP) to approximate the mean($\mu$) and the log variance($\phi$) of the latent distribution. The latent variable $z$ is then reconstructed deterministically through the reparameterization trick~\cite{kingma_auto-encoding_2014} i.e. $z = \mu + \sigma\epsilon$, where $\sigma = e^{\frac{1}{2} \phi}$, $\epsilon$ is an auxiliary noise variable sampled from standard Gaussian such that $\epsilon \sim \mathcal{N}(0, 1)$. This technique makes the stochastic estimation of the latent variable $z$ differentiable. The decoder reconstructs $R \sim A^{-1}$ from $z$ such that $\norm{I - R^{T} A R}_F \approx 0$. We implemented the same optimisation function as the traditional SPAI algorithms, because the eigenvalue decomposition operation is not differentiable and this least-square optimisation implicitly reduces the condition number of $AR$ which is what we are interested in, as the condition number of an identity matrix and its multiplications with an arbitrary scalar is 1. The sparsity pattern of $R$ is constrained by a mask $M$ which has a similar sparsity pattern of $A$ at the output layer. Without the prior knowledge of $A^{-1}$, this is the simplest method of predefining the sparsity pattern as stated in \cite{wathen_preconditioning_2015}. Furthermore, we allow a small percentage of extra non-zeros in addition to the existing non-zero entries, which we have found to significantly improve the model performance in our experiments.

\begin{figure}
\includegraphics[width=\textwidth]{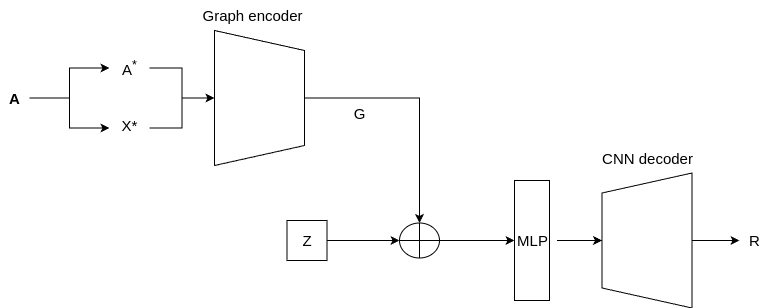}
\caption{Schematic diagram of the inference process} 
\label{fig:Model inference diagram}
\end{figure}

For inference, as shown in Figure~\ref{fig:Model inference diagram} we pass the adjacency matrix $A^{*}$ and the node feature $X^{*}$ of an unseen stiffness matrix $A$ from the same family of linear PDE problems through the graph encoder to generate the conditional information $G$, then concatenate with $z$ sampled from $z \sim \mathcal{N}(\mu,\sigma^2)$ and pass it through the decoder to generate the preconditioner $R$. Due to the nature of our model, multiple different $R$ can be generated for a single $A$.

\subsection{Loss function}
\label{subsec:loss function}
In general, the family of VAE networks optimise the well-known evidence lower bound (ELBO)~\cite{kingma_auto-encoding_2014}:

\begin{equation}
L = \mathbb{E}[\log p( R | Z )] - \alpha D_{KL} [q( Z | X ) || p( Z )],
\end{equation}
where the prior $Z \sim \mathcal{N}(0,I)$
s.t. $p(Z) =\prod p(z_i) = \prod \mathcal{N}(z_i | 0,I)$.
This type of loss function contains a reconstruction term, which is the first term that calculates the expected negative reconstruction error. The second term is the KL-divergence term, and it can be regarded as a regulariser which encourages the posterior distribution to be close to the prior, in this case the prior is assumed as the standard Gaussian. In our case, we replace the reconstruction term with $\norm{I - R^{T} A R}_F$ and minimise the following loss function:

\begin{equation}
\label{eq:vanilla loss function}
\begin{split}
%L =\mathbb{E}[\norm{I - A (M \odot R)}_F^2] + \alpha D_{KL} [q_{MLP} ( Z | X ) || p( Z )]\\
    L =\mathbb{E}[\norm{I - R^{T} A R}_F^2] - \alpha D_{KL} [q ( Z | X ) || p( Z )]\\
    \textbf{\textit{for}} \quad X = GNN_\theta ( A^* ,X^*) \oplus CNN_\theta ( A^{-1}) \;,
\end{split}
\end{equation}
where $\alpha$ is the regularising parameter, which is chosen to be 0.1. ${GNN}_\theta (\cdotp)$ and ${CNN}_\theta (\cdotp)$ represent the GNN encoder and CNN encoder as shown in Figure~\ref{fig:Model schematic diagram}, and $\oplus$ is the aggregation operator.

\section{Experiments}
\label{sec:experiments}
In the first set of computational experiments that we consider (subsections~\ref{subsec:Poisson} and~\ref{subsec:Poisson's result}) the dataset, ${\cal A}$, is drawn from piecewise linear finite element discretizations of a family of second order elliptic PDEs in two dimensions. The second set of experiments considers a more challenging dataset, ${\cal A}$, that is drawn from discontinuous (piecewise quadratic) Galerkin discretizations of a family of two-dimensional
biharmonic problems (subsections~\ref{subsec:Biharmonic} and~\ref{subsec:Biharmonic result}). Such problems are known to lead to highly ill-conditioned matrix systems upon discretization (as illustrated below). Both experiments are trained on a single NVDIA RTX 3090 GPU with 24 Gigabyte of VRAM. The dataset preparation of both problems takes less than 30 mins for the largest problem size. For the largest problem size of 2D Poisson's problem $1873 \times 1873$, the training converges at 200 epochs and around 6 mins per epoch. Similarly, for the largest biharmonic problem, of size $1089 \times 1089$, the training converges at 260 epochs and around 2 mins per epoch. The training cost is expected to grow linearly with the total number of hyperparameters, which depends on many different factors, such as the discretized mesh density, parallelisation, depth of the model, number of the channels of the CNN layers, etc. For practical implementation of the model, fine tuning of the hyperparameters is required to find the optimal balance between the training cost and model performance, though this is not the focus of this paper.

\subsection{2D Poisson's Problem}
\label{subsec:Poisson}

We generate sets of $N=2000$ matrices based upon the piecewise linear finite element discretization of PDEs of the form
\begin{equation}
    \underline{\nabla} \cdot (f(\underline{x}) \underline{\nabla} u) = g(\underline{x}),
\end{equation}
in two dimensions. For simplicity we consider a unit square domain and impose Dirichlet conditions on the entire boundary. The dimension, $n$, of the resulting stiffness matrices, $A$, is equal to the number of interior node points in the unstructured triangular mesh, whilst the entries of $A$ depend upon the node locations and the choice of $f(\underline{x})$, which is drawn from a family of polynomial functions that are positive on the unit square. Each set of $2000$ matrices is generated on the same mesh and is randomly split into 1600 training samples and 400 for testing.

\subsection{Results and discussion - Poisson family}
\label{subsec:Poisson's result}

This section demonstrates the performance of the preconditioners generated by our GCVAE model when applied with a CG solver using a relative convergence criterion of 1.0e-5. Included within our results is a comparison with two baseline methods: Jacobi preconditioning and Super Nodal incomplete LU factorisation (SPILU)~\cite{li_splu05} (using its symmetric mode to obtain IC preconditioning). Note that the performance of the latter comparator depends critically on the value of a ``drop tolerance'' parameter, which controls the amount of fill-in that is permitted during the incomplete factorization of $A$. When this is very small the IC preconditioner is highly effective but at the expense of significant additional computational cost (reducing the overall efficiency of the solver); when the drop-tolerance is larger the cost of computing and applying the preconditioner goes down but at the expense of a much larger number of CG iterations. 

\begin{figure}
\includegraphics[width=0.5\textwidth]{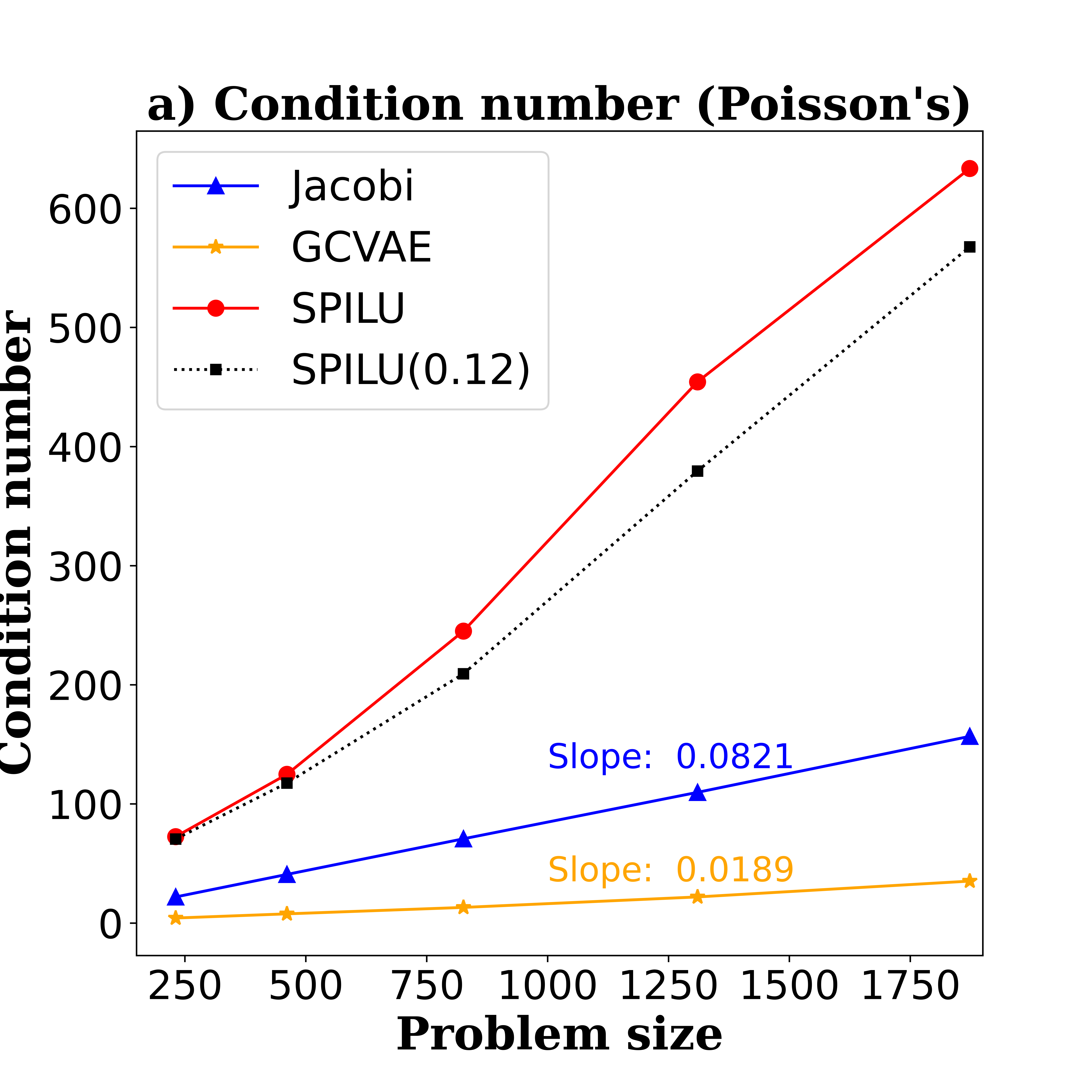}
% \hspace*{1mm}
\includegraphics[width=0.5\textwidth]{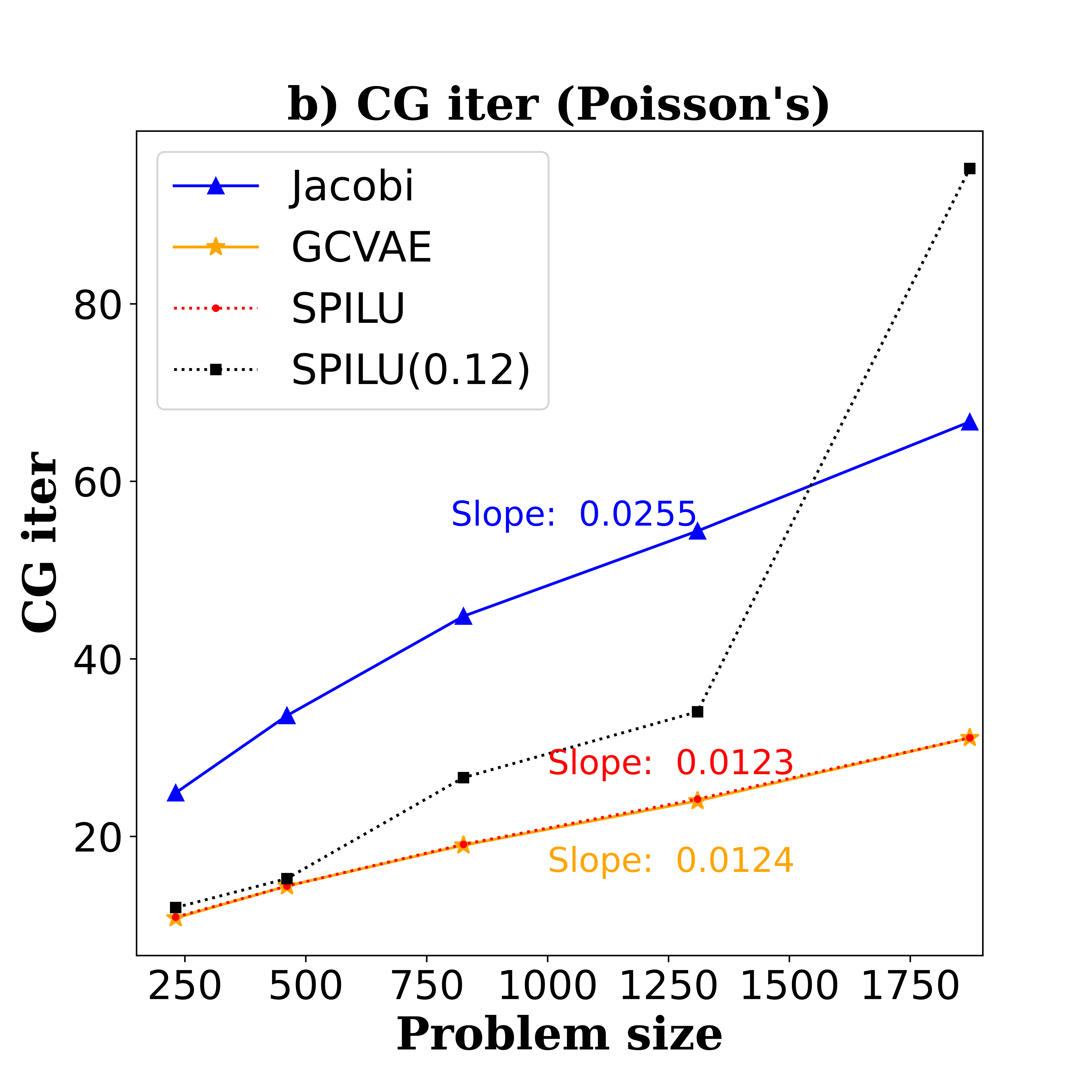}
\caption{Comparison of the condition number and CG algorithm iteration count for Jacobi preconditioning, SPILU preconditioning and our proposed method (GCVAE) when applied to discretizations of a family of second order operators} 
\label{fig:GCVAE vs Jacobi}
\end{figure}

Figure~\ref{fig:GCVAE vs Jacobi} (left) shows the average estimated condition numbers for preconditioned systems with each choice of preconditioner. For the SPILU case we have artificially selected the drop-tolerance for each problem size so as to match the number of iterations taken using the GCVAE preconditioner. Consequently, by design, the equivalent curve in Figure~\ref{fig:GCVAE vs Jacobi} (right), which shows average CG iterations, completely overlays the GCVAE curve. Also in this graph are the average iteration counts for SPILU applied in its more conventional form, with a constant drop-tolerance (in this case $0.12$). It is clear from these examples that both the condition number and the CG iteration counts with Jacobi preconditioning grow at a much faster rate than for the GCVAE approach, demonstrating that our model will out-perform the Jacobi method on total execution time for sufficiently large problems.

Comparison against SPILU is less straightforward due to the trade-offs in the choice of drop-tolerance described above. The SPILU algorithm uses $(A^T + A)$ based column permutation \cite{superlu_ug99} which causes the condition number to increase rapidly with the problem size as shown in Figure~\ref{fig:GCVAE vs Jacobi} (left). This illustrates the limitation of using the condition number as the measure of quality, which is why we prefer to use iteration count. Nevertheless, even with this measure, Figure~\ref{fig:GCVAE vs Jacobi} (right) shows that the parameters of the SPILU method need to be carefully tuned to achieve optimal performance. By carefully reducing drop-tolerance as $n$ increases we are able to match the iteration counts of the GCVAE preconditioner, though the latter has significantly lower condition numbers. In Figure~\ref{fig:density comparison} (left) we also compare the density of non-zero entries in these two preconditioners for different choices of $n$, in order to give an indication of their computational costs. It can be seen that for sufficiently large systems the GCVAE preconditioner will be expected to have fewer non-zeros. Furthermore, to apply the SPILU preconditioner requires backward and forward substitution to be applied which cannot naturally be done in parallel, whereas the SPAI preconditioning is ideally suited to parallel implementation (since it just requires a sparse matrix-vector multiplication). The execution time benchmark test has not been carried out as our model is currently implemented in a non-optimised manner in a Python environment, whereas the state-of-art numerical software SPILU is implemented in a highly optimised C-language environment. Nevertheless, the advantage of the data driven approach is that, once it is tuned and trained, it can be used as a black box tool with execution time complexity close to $\mathcal{O}(n)$.

\subsection{Biharmonic Problem}
\label{subsec:Biharmonic}
We again generate sets of $N=2000$ matrices, this time based upon the piecewise quadratic discontinuous Galerkin discretization of fourth order PDEs of the form
\begin{equation}
    {\nabla}^2 (f(\underline{x}) {\nabla}^2 u) = g(\underline{x})
\end{equation}
in two dimensions. We consider a unit square domain and impose Dirichlet conditions on both $u$ and $\nabla^2 u$ on the entire boundary. For a given triangular mesh, the dimension, $n$, of the resulting stiffness matrices, $A$, is much greater than for the piecewise linear approximations previously considered and the condition number of the matrix $A$ is much larger (hence this is a considerably more challenging test problem). The individual non-zero entries of $A$ depend upon the mesh node locations and the choice of $f(\underline{x})$, which is again drawn from a family of polynomial functions that are positive on the unit square. As previously, each set of $2000$ matrices is generated on the same mesh and is split into 1600 training samples and 400 for testing. For this problem, we allow 20\% extra number of non-zeros upon the existing non-zero entries during training and designed 4 CNN layers for its encoder and decoder to guarantee the convergence and model performance. In contrast, we only used 3 CNN layers with less channels per layer for 2D Poisson's problem.

\subsection{Results and discussion - Biharmonic family}
\label{subsec:Biharmonic result}

\begin{figure}
\includegraphics[width=0.5\textwidth]{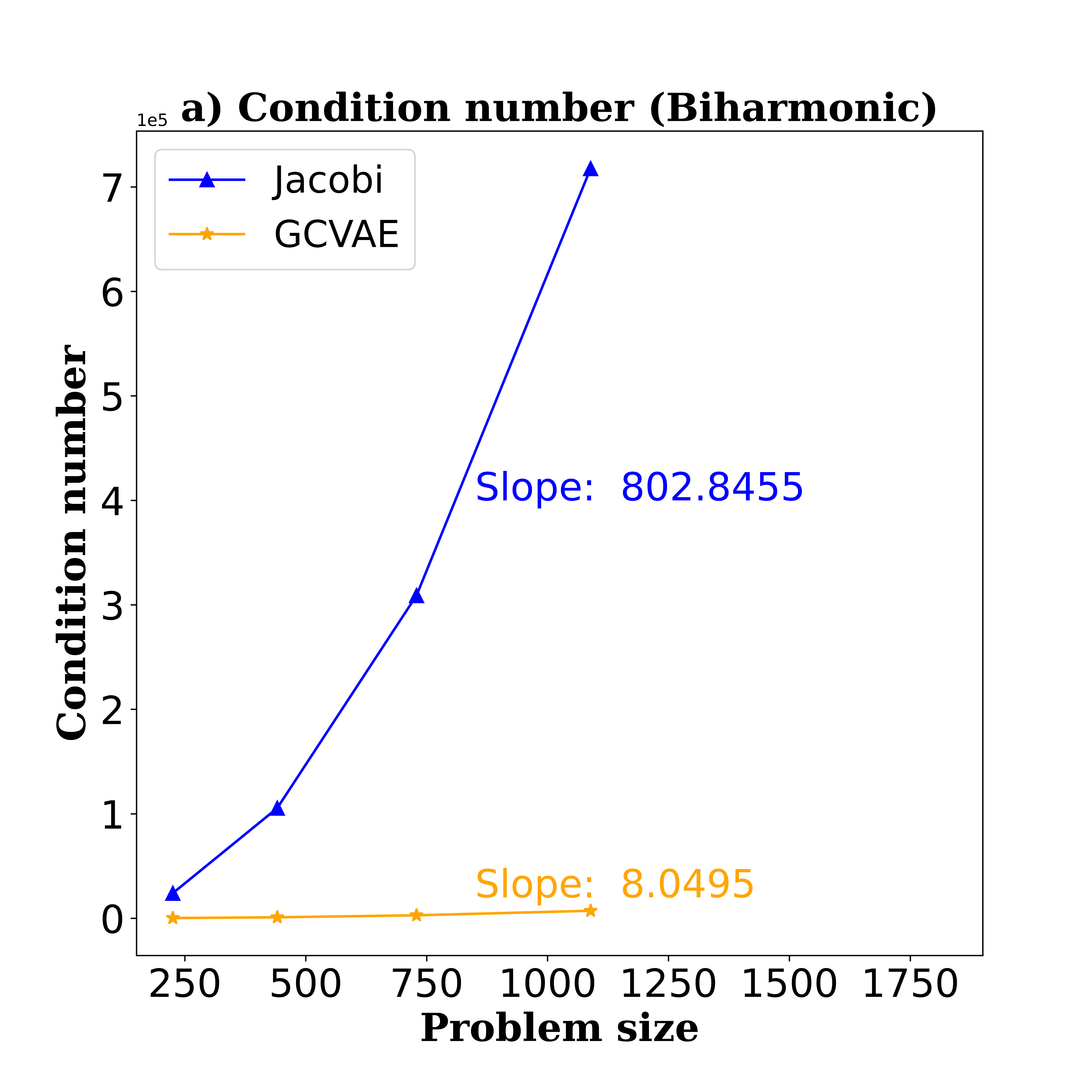}
% \hspace*{1.5mm}
\includegraphics[width=0.5\textwidth]{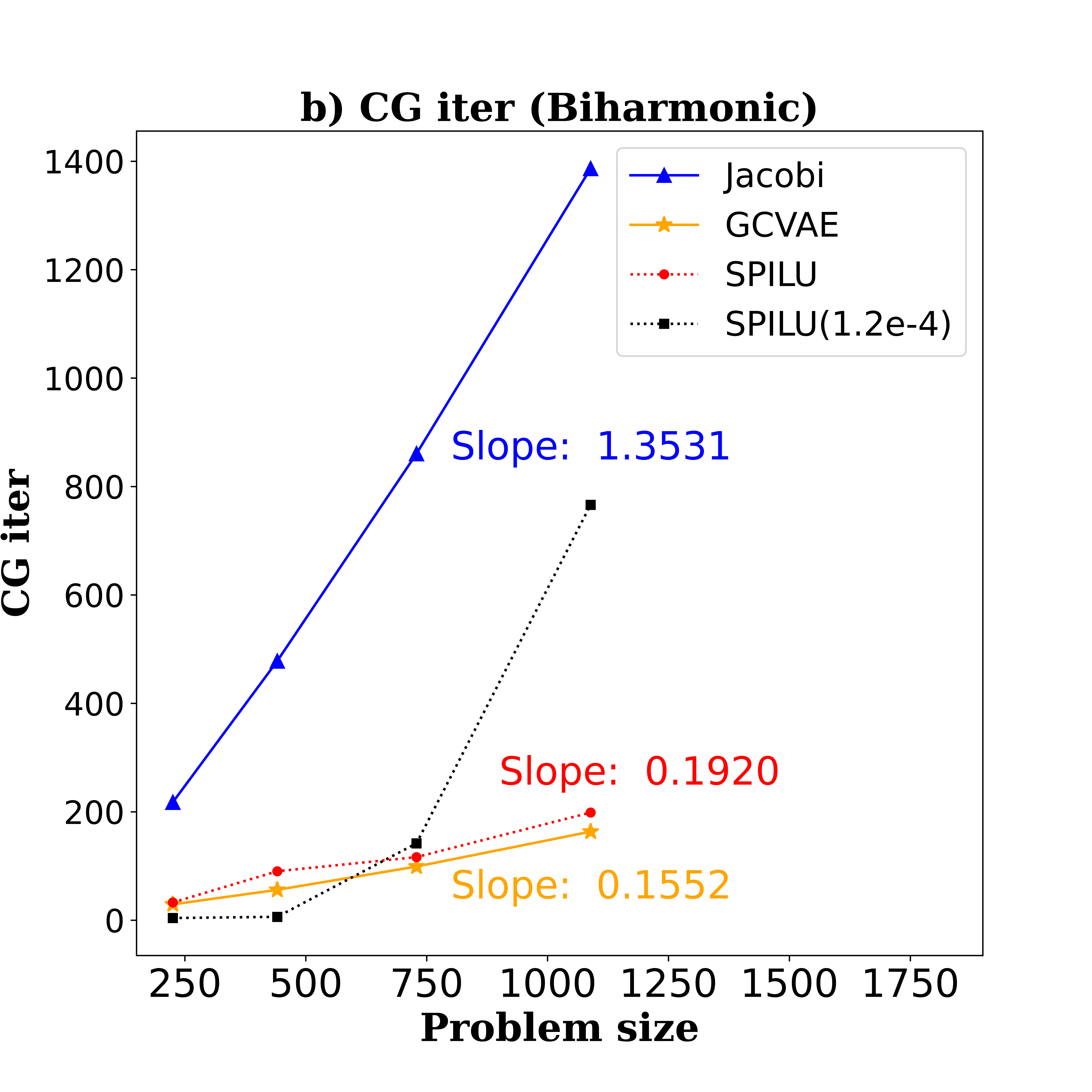}
\caption{Comparison of the condition number (in the unit of $1.0 \times 10^{5}$) and CG algorithm iteration count for Jacobi preconditioning, SPILU preconditioning and our proposed method (GCVAE) when applied to discretizations of a family of fourth order operators. The condition numbers of the SPILU approaches are not illustrated in \textbf{a)} as they are too large to fit into the plot.} 
\label{fig:biharmonic iteration comparison}
\end{figure}

Similarly, in this subsection, we compare our method against two baseline methods: Jacobi and SPILU. As shown in Figure~\ref{fig:biharmonic iteration comparison}, for this ill-conditioned problem, as the problem size increases our method significantly outperforms Jacobi preconditioning in terms of both the condition number reduction and the CG convergence rate.

\begin{figure}
\includegraphics[width=0.5\textwidth]{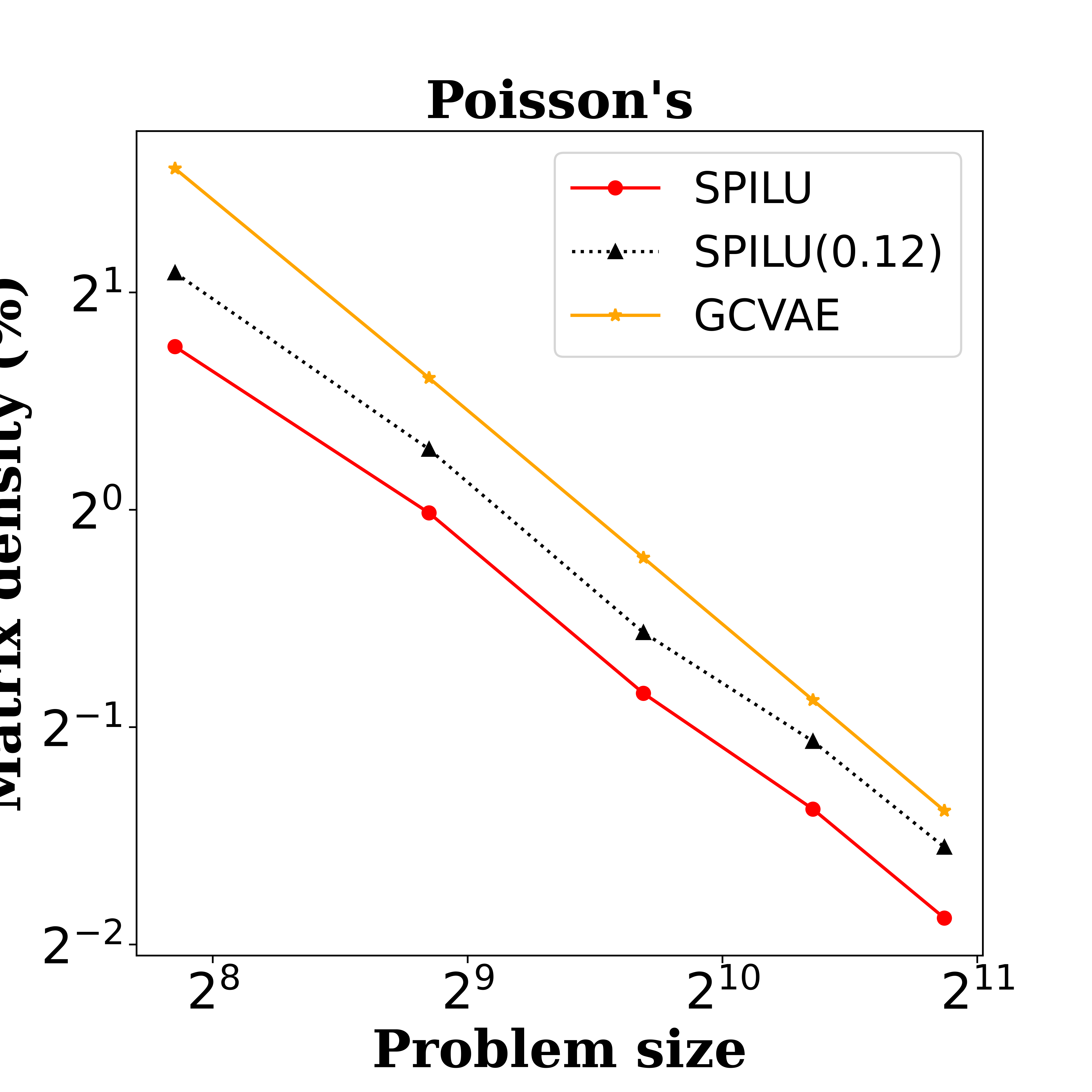}
\hspace*{1.5mm}
\includegraphics[width=0.488\textwidth]{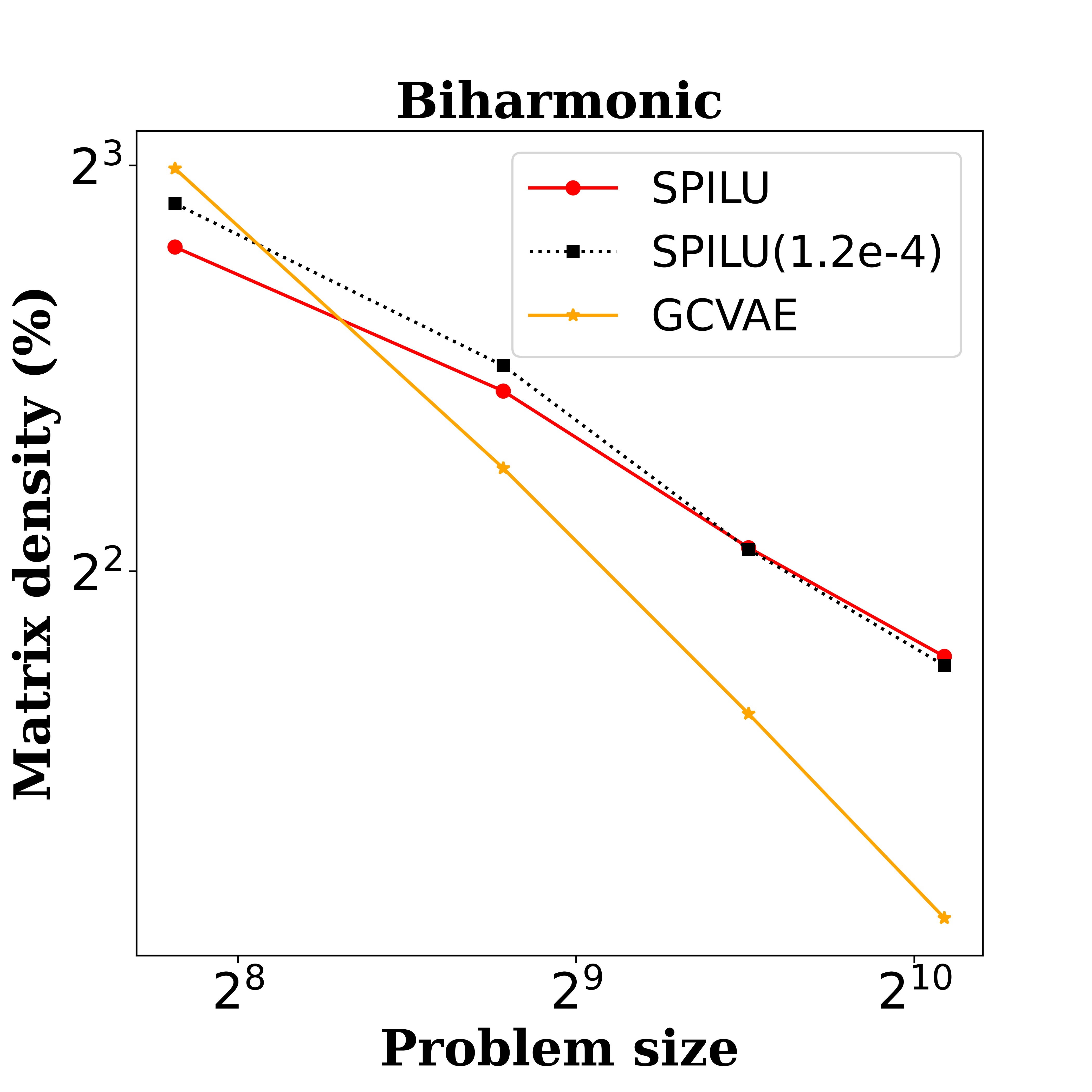}
\caption{Comparison of the proportion of non-zero entries in the preconditioners generated via the SPILU and GCVAE methods for the second order (left) and fourth order (right) problems} 
\label{fig:density comparison}
\end{figure}

Comparison against SPILU shows similar features as for the previous test case. For the reasons described previously we do not find condition number to be a useful metric in this case and therefore focus on iteration counts. Figure~\ref{fig:biharmonic iteration comparison} (right) shows that it is possible to tune the value of drop-tolerance for each problem size in order to match the number of CG iterations obtained with the GCVAE preconditioner, as well as showing the growth in iterations when a constant drop-tolerance is used (in this case $1.2 \times 10^{-4}$). Equally significantly however, we see from Figure~\ref{fig:density comparison} (right) that the number of non-zeros required for the GCVAE preconditioner is significantly fewer than for the SPILU preconditioner, even when drop-tolerance is tuned to match the number of iterations. Combined with the fact that application of SPILU requires backward and forward substitution, we observe that this preconditioner is much more expensive to apply than our proposed approach. 

\section{Conclusions and Future Work}
\label{sec:conclusion}

\subsection{Conclusions}
This paper proposes a novel generative modelling framework to generate sparse approximate inverse preconditioners for matrix systems arising from the mesh-based discretization of families of self-adjoint elliptic partial differential equations. The approach has shown excellent potential in terms of its ability to reduce the condition number of the systems considered and to increase the convergence rate of the conjugate gradient solver. This performance has been demonstrated for two different families of partial differential equations: of second and fourth order respectively. Furthermore, comparing against a state-of-art factorisation-based preconditioning technique we observe that, for sufficiently large systems, our proposed approach should require fewer floating point operations to apply and that it is better suited to future parallel implementation. 

\subsection{Limitation and Future Work}
Based upon the evidence presented here we believe that the proposed GCVAE approach has significant potential. However, there still exists several limitations which will need to be considered for future development. The most significant of these is that our training phase currently requires the inverse matrices $A^{-1}$ to be known for our training set, even though the dataset preparation cost is negligible compared with the training time for the experimented problem size, this cost will quickly grow to an unaccepted level when generalising to solve real world problems. Consequently, we propose to investigate ways to relax this restriction (e.g. through the use of training data from coarser finite element grids or the application of techniques from algebraic multigrid to obtain coarser representations of training matrices and inverting these). Furthermore, despite our current masking scheme allowing flexible mask selection that can be tuned to deal with more complex PDE problems, the sparsity pattern does need to be decided prior to training which may restrict the generalisation capability of the model. A learning-based masking approach could be more flexible and efficient, so will be attempted in further development. Finally, the work in this paper is limited to self-adjoint elliptic problem, which result in symmetric positive-definite stiffness matrices. We also intend to generalise our model to other families of PDE problems, which lead to non-symmetric and/or indefinite linear systems.

\begin{credits}
\subsubsection{\ackname} The first author gratefully acknowledges receipt of a doctoral training award from the UK Engineering and Physical Sciences Research Council. 

\subsubsection{\discintname}
The authors have no competing interests to declare that are
relevant to the content of this article.
\end{credits}

\bibliographystyle{splncs04}
\bibliography{references/GCVAE}

\end{document}